\RequirePackage[2020-02-02]{latexrelease}
\documentclass[10pt,letterpaper]{article}
\usepackage[top=0.85in,left=2.75in,footskip=0.75in]{geometry}

\usepackage{amsmath,amssymb}

\usepackage{changepage}

\usepackage[utf8x]{inputenc}

\usepackage{textcomp,marvosym}

\usepackage{cite}

\usepackage{nameref,hyperref}

\usepackage[right]{lineno}

\usepackage{setspace}
\usepackage{microtype}
\DisableLigatures[f]{encoding = *, family = * }

\usepackage[table]{xcolor}

\usepackage{array}
\usepackage[printwatermark]{xwatermark}
\newcolumntype{+}{!{\vrule width 2pt}}

\newlength\savedwidth

\usepackage[aboveskip=1pt,labelfont=bf,labelsep=period,justification=raggedright,singlelinecheck=off]{caption}

\usepackage{csquotes}

\makeatletter
\renewcommand{\@biblabel}[1]{\quad#1.}
\makeatother

\usepackage{lastpage,fancyhdr,graphicx}
\usepackage{epstopdf}
\fancyheadoffset[L]{2.25in}
\fancyfootoffset[L]{2.25in}
\begin{document}

\vspace*{0.2in}

\begin{flushleft}
{\Large
\textbf\newline{Statistical Learning to Operationalize a Domain Agnostic Data Quality Scoring} 
}

Sezal Chug\textsuperscript{1},
Priya Kaushal\textsuperscript{1},
Ponnurangam Kumaraguru\textsuperscript{1},
Tavpritesh Sethi\textsuperscript{1}
\\
\bigskip
\textbf{1} Department of Computer Science, Indraprastha Institute of Information Technology, New Delhi, Delhi, India, 110020
\\
\bigskip

\textcurrency Current Address: Department of Computer Science, Indraprastha Institute of Information Technology, New Delhi, Delhi, India, 110020 

* sezal17101@iiitd.ac.in
\\
* priya17081@iiitd.ac.in
\\
* pk@iiitd.ac.in
\\
* tavpriteshsethi@iiitd.ac.in

\end{flushleft}

\section*{Abstract}\label{abstract}
\subsection*{Background}
Data is expanding at an unimaginable rate, and with this development comes the responsibility of the quality of data. Data Quality refers to the relevance of the information present and helps in various operations like decision making and planning in a particular organization. Mostly data quality is measured on an ad-hoc basis, and hence none of the developed concepts provide any practical application. 
\subsection*{Methods}
The current empirical study was undertaken to formulate a concrete automated data quality platform to assess the quality of incoming dataset and generate a quality label, score and comprehensive report. We utilize various datasets from healthdata.gov, opendata.nhs and Demographics and Health Surveys (DHS) Program to observe the variations in the quality score and formulate a label using Principal Component Analysis(PCA). The results of the current empirical study revealed a metric that encompasses nine quality \enquote*{ingredients}, namely provenance, dataset characteristics, uniformity, metadata coupling, percentage of missing cells and duplicate rows, skewness of data, the ratio of inconsistencies of categorical columns, and correlation between these attributes. The study also provides a case study and validation of the metric following Mutation Testing approaches.  
\subsection*{Findings}
The proposed system quantifies the provided data and evaluates them at subjective and objective levels. Eventually, it formulates a metric which encompasses the given ingredients and judges the quality of any incoming dataset. Using a case study of DHS Indian dataset, the study illustrates the use of the proposed metric. The case study following DHS India dataset shows that DQ scores range between 85-93\% with a gradient increase among the years. Mutation testing approach provides empirical proof that for any dataset, if some percentage of noise is inserted or removed, variations in data quality vary accordingly. 
\subsection*{Interpretation}
Due to the growing technology upgradations in data collection and processing, there is a constant gradient increase in the quality of data. With the modified Mutation Testing approach, the research validates the metric and completely captures the essence of any incoming dataset. This research study provides an automated platform which takes an incoming dataset and metadata to provide the DQ score, report and label. The results of this study would be useful to data scientists as the value of this quality label would instill confidence before deploying the data for his/her respective practical application.
\subsection*{Funding}
The following research study has been funded by the Population Council under the umbrella of ICMR. We also acknowledge the valuable suggestions provided by ICMR, NDQF and Population council while preparing the paper.

\noindent \textbf{Keywords.} Data Quality Framework,Quality Metric, Demographics and Health Surveys (DHS) Program, Dataset Quality Label, Nutrition Label Approach, Metadata Coupling, Mutation Testing

\section{Introduction}\label{introduction}
With advancements in technology, use of data has become immensely influential and hence the importance of its quality. Amount of data created and consumed globally is forecast to increase rapidly, from 64.2 zettabytes in 2020 to more than 180 zettabytes in 2025 \cite{datagrowth}. Data is anticipated to become an integral part of our lives and day-to-day functionality. While human-generated data is experiencing an exponential growth rate, machine data is increasing even more rapidly \cite{dataisincreasing}. This ever-increasing speed of data growth has introduced several challenges encompassing high operational costs at data processing stations for storage and processing \cite{chanllengesofdata}. Evaluating data based on subjective and objective levels has become the need of the hour. 

Data quality refers to the relevance of information for its use in a particular application. Low data quality has been curbing the growth of various organizations by preventing them from performing to their full potential \cite{intro1}. Analyzing the data quality levels can help organizations identify the pitfalls that need to be resolved to enhance their clarity. Furthermore, inaccurate data can be identified and fixed to ensure that executives, data analysts, and other end users work with accurate and efficient information. 

With the advent of the machine learning/artificial intelligence realm, quality of data has mainly been neglected under the assumption that the data feeding these algorithms is of high quality. There is always more focus on learning algorithms and models instead of ensuring data quality. A study conducted by Wand et. al. \cite{2} stated that the actual use of the data is outside of a researcher's control, however, it is essential to provide data conforming to a particular level to ensure its proper usage. Poor data quality can have a severe impact on the overall effectiveness of data in an organization. The \enquote{new normal} of massive generation, utilization and elimination of data has urged researchers to consider its data quality aspect. 

Essential elements of good data quality include \textit{Completeness, Consistency, Concordance, and Conformity} to the standard data formats created by a particular organization \cite{datadimensions}. Meeting all of the above mentioned factors is necessary to ensure that datasets are reliable, trustworthy, and suitable for use. It is difficult to define quality for data because unlike manufactured products, data do not have physical characteristics that allow quality to be easily assessed \cite{1}. The actual use of the data is outside of the designer's control, hence a design-oriented definition of data quality is necessary \cite{2}.

Currently, most data quality measures provide an ad hoc basis to solve specific problems, as suggested by Huang et al. \cite{Huang} and Laudon \cite{Laudon}. Pipino et al. \cite{Pipino} proposed a new data management paradigm to help unify the diverse efforts using a flexible schema that pursued data integration and unification. Starting with a pretty high-level description of the components of data quality by Veregin \cite{1}, researchers have gone into intricate aspects of datasets. Wand et. al. \cite{2} in 1996 introduced us to data quality dimensions in ontological foundations, which have now become advanced and more detailed. Jayawardene et al. \cite{3} in 2015 provided comprehensive classifications of data quality dimensions, which helped develop a streamlined and unified set of quality dimensions. Initially, researchers gave an idea about quality dimensions which encompass the \textit{accuracy, precision, consistency, and completeness} of an incoming dataset. Further, a few researchers elaborated on these principles and established quantities for these DQ dimensions which help elucidate the perception of data quality. \textit{reliability, null values, size of data, correctness, accuracy, conformance and duplication}, were a few core dimensions central to the practical analysis of formal data quality requirements. 

Some of the dimensions discussed by the above mentioned researchers are given below.
\begin{itemize}
    \item Provenance provides useful information on the history of a dataset, such as when it was last updated, the source data and the authority that certifies the dataset, if any \cite{10}.
    \item Uniformity refers to whether instances of data are either stored, exchanged, or presented in a format that is consistent with the domain of values and consistent with other similar attribute values \cite{3}.
    \item Accuracy is the first and foremost requirement that many users expect from data. The accuracy of an entire database can be measured by finding the fraction of incorrect tuples in the database \cite{8}.
    \item Missing values refer to a null value as a missing value \cite{3}.
    \item Duplication of rows has been defined as a measure of unwanted duplication existing within or across systems for a particular field, record, or data set \cite{6}. 
\end{itemize}

In another research by Sun et al. \cite{Sun}, metadata matching with the purpose of information integration was considered to be an essential aspect of the assessment of data quality. Data quality documentation plays a key role in many standards as data manuals are essential for effective use. Pantulkar and Srinivas \cite{Srinivas} insisted that semantic similarity has a vital role in natural language processing and application. He proposed three different semantic similarity approaches in their research, i.e. cosine similarity, path-based approach and feature-based approach. This approach aided in performing tagging and lemmatization to  calculate the metadata matching score. Many previous works of literature have given a brief idea about these dimensions that define the quality of data; however, none of them partakes the importance of metadata into their quality judgement. Holland et al. \cite{Holland} proposed method which sets a nutrition label approach by using \enquote{ground truth} data as a comparison dataset for quality judgment. 

The previous research conducted insinuates fundamental principles which are necessary for developing data quality parameters but incomplete to formulate usable metrics. More studies need to be conducted on how to operationalize these formally derived dimensions in design practice. As famously said by Willcocks and Lester \cite{WILLCOCKS}, “what gets measured gets managed”, hence quality labels measuring the dimensions of quality of data signify a crucial management element in the domain of data quality. However, the emergence of new classifications of data lack a shared understanding amongst its various organizations for a widely accepted practice for calculating data quality scores. Hence, research needs to be conducted to bridge this gap of quality. 

On the similar lines, the ibid empirical study aims to formalize some data quality dimensions and suggests a nutrition label approach towards building a quality label that captures the data quality of any incoming dataset and evaluates it on these quality ingredients to generate a DQ score. This widely accepted quality metric would quantify data and measure the degree to which it fits our purpose. This study defines certain factors which can be used on any given dataset to measure its quality.

The current study undertakes a path by defining quality indicators called \enquote*{data quality ingredients} that incorporates a semantic perspective on data quality \cite{3}. After a thorough analysis, the current research came to conclude \enquote{ingredients} of a dataset that will further adjudge data quality. These nine quality \enquote{ingredients} are \textbf{provenance, dataset characteristics, uniformity, metadata coupling, percentage of missing cells and duplicate rows, skewness of data, the ratio of inconsistencies of categorical columns, and correlation between these attributes}. These \enquote{ingredients} would be guiding factors that help calculate the final score that would help achieve better data quality. The study also presents a comprehensive report which gives an overview of the \enquote{ingredients} of the dataset and suggests ways and means to improve this quality score. Further, we present a streamlined automated platform that is instrumental in the pursuit of measuring data quality. Setting up an empirical metric to evaluate data quality leads to increased profitability, which helps make better decisions for organizations by improving data usability and reducing storage and processing costs.

In the further sections of this study we will describe these dimensions of data quality in detail, elaborating upon the platform that generates the quality label, DQ score and the comprehensive report. It would also include the validation and application of the proposed model which helps instill faith towards our approach.

\section{Dataset retrieval}\label{dataset}
The Demographics and Health Surveys (DHS) Program \cite{dhs} dataset of India contains restricted survey data files for legitimate academic research. We collected healthcare survey data from the DHS website for countries Myanmar, Ethiopia, Zimbabwe, Maldives, Nepal, Nigeria, Afghanistan, Bangladesh and Cambodia for the year 2015-16 for training purposes. Along with these countries, healthcare datasets from HealthData.gov (https://healthdata.gov/) and Health and Social Care Open Data platform(opendata.nhs) were also combined to finalize 200 Training datasets. DHS Indian datasets over the years 1992-93, 1998-99, 2005-06 and 2015-16 were collected for testing the formulated metric. Data quality assessment of the collected data is conducted by dividing the dataset into sections based on the DHS Recode Manual and individually analysing each section. The formulated dataset is attached in the references \cite{dataset}.

The numerous takeaways of the proposed model are discussed in further sections, which suggest a data quality metric to adjudge the quality of data and provide empirical validation to the proposed metric. 

\section{Methods}\label{methods}
\begin{figure*}
        \centering
        \includegraphics[height=7cm, width=6in]{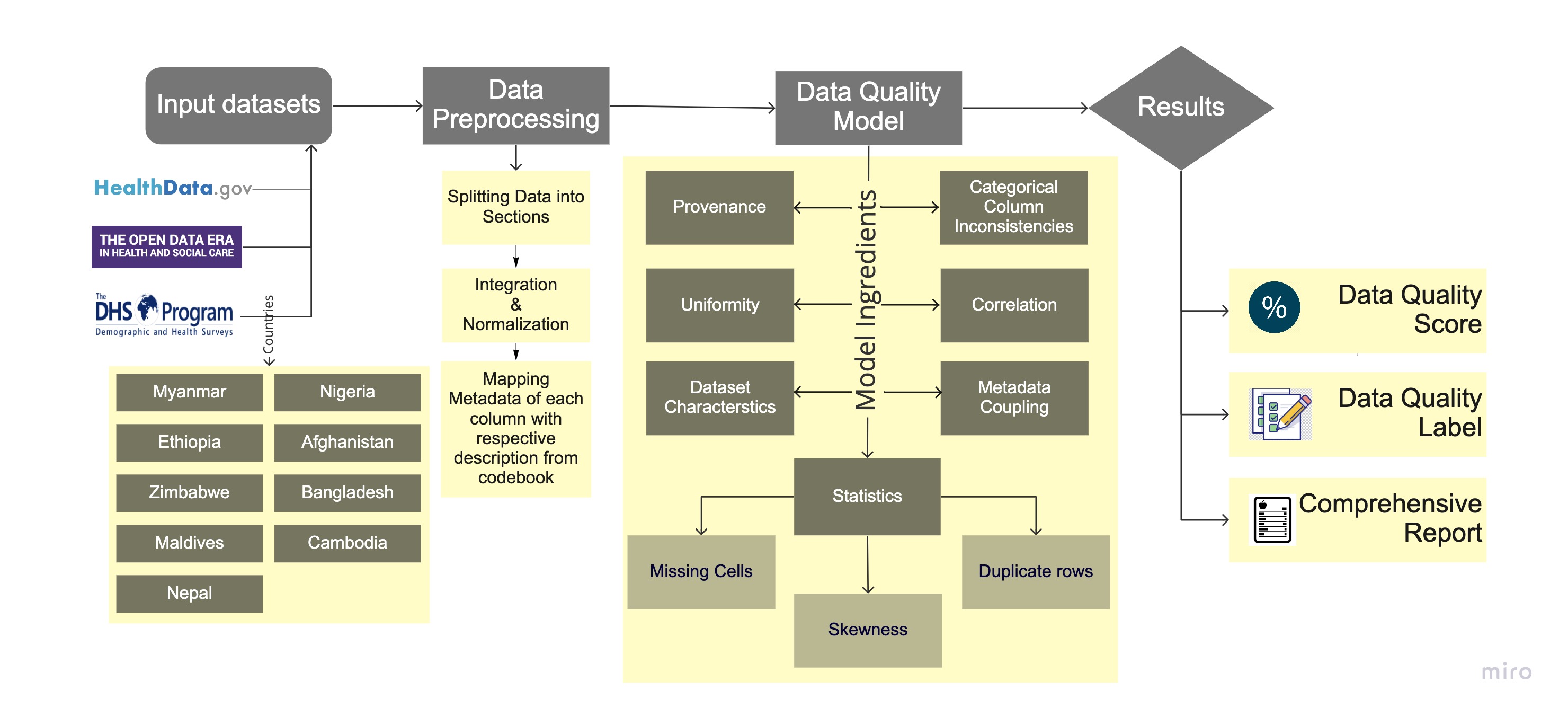}
        \caption{Research Methodology Diagram}
        \label{fig: methodology}
    \end{figure*}

The research methodology diagram illustrates the formulation of quality label and data quality metric using training datasets. As shown in the figure \ref{fig: methodology}, we calculate the values of nine DQ ingredients of our model for all the training datasets to formulate a sheet containing datasets as rows and ingredients as columns. After preprocessing, normalization and quantification, this combined dataset is utilized to formulate the required metric. The study finalized \textit{ nine {\lq{ingredients}\rq}}, as mentioned in the section \ref{introduction} are used to judge every incoming dataset to generate a nutrition label, data quality score and comprehensive report detailing these factors.

\subsection{Data Quality Ingredients}\label{ingredients}
   \begin{figure}
        \centering
        \includegraphics[height=8cm, width=8cm]{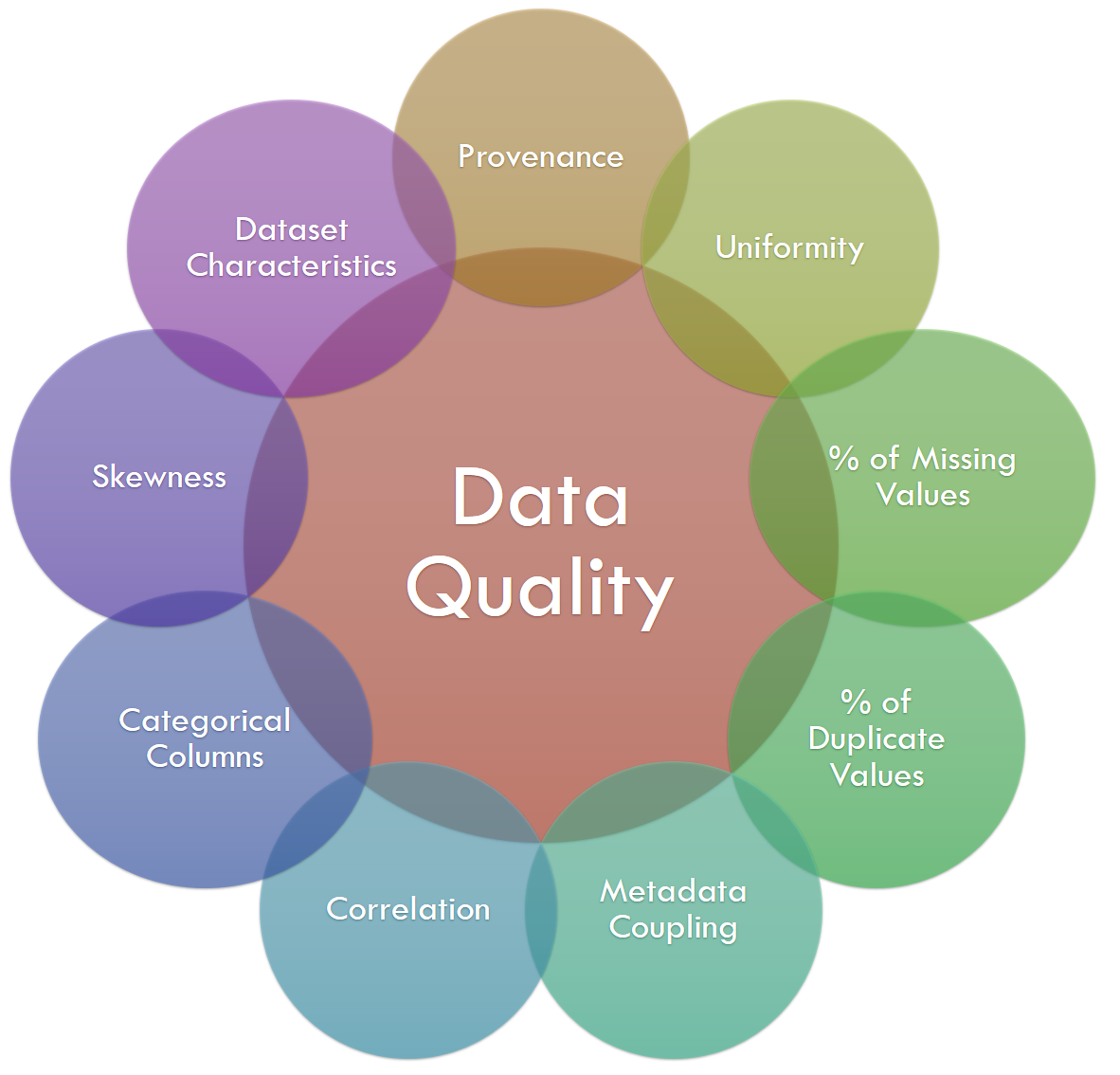}
        \caption{Data Quality Ingredients focused upon in the study}
        \label{fig: ingredients}
    \end{figure}  
The below subsections contain a detailed description of the dimensions of data quality of the current study as illustrated in the figure \ref{fig: ingredients}.

\subsubsection{Provenance}\label{provenance}
    Reliability indicates whether the information can be counted on to convey the right information to a researcher \cite{2}. Taking inspiration from the above dimension, the study starts with \textbf{Provenance}. It refers to the record trail, which accounts for the origin of the dataset and details about the latest version. It provides baseline information for assessing authenticity, integrity and helps in enabling the trust of the researcher for using the dataset. The provenance can be part of the local repository, present on the website or may be present in a separate file \cite{p1}. The parameters used for calculating the provenance of a dataset are origin/source, author, latest updated date and data accessibility. All the parameters are given equal weightage to calculate the provenance percentage of the dataset. It helps to answer questions like “who”, “how”, “where”, “when”, and “by whom” was data produced. This is also referred to as data lineage, which includes either authentic government sources which are given full score for origin or private sources like kaggle or opendata where this score is decided on the basis of the number of usages of the dataset. The number of years between last updated date and present give us the percentage of weightage for correctness. Data accessibility includes permissions for intellectual property, the format of the dataset, the kind of preprocessing performed \cite{p2} are also calculated while measuring provenance. 
\subsubsection{Dataset Characteristics}\label{Dataset Characteristics}
    After the data is imported from the original path, preliminary checks are conducted, and the values are cross-checked with the data provided on the author's website. Accuracy is the first and foremost requirement that many users expect from data. Hence, data is correct if it conveys the same meaning lexically, syntactically and semantically \cite{3}. The \textbf{Dataset Characteristics} ingredient includes the comparison of mean, median, mode, standard deviation, range of values (Min-Max values) and the total number of observations in a dataset. These values are then matched with the values scrapped of the source website to generate a percentage of correctness in these characteristic quantifications \cite{10}. All parameters are given equal weightage to calculate the final percentage, which helps judge the quality of data at a discrete level and provide the user with a high level of information about the dataset.
\subsubsection{Uniformity}\label{Uniformity}    
    \textbf{Uniformity} in a datasets highlights instances within a dataset which are consistent with values present on the sources of the dataset \cite{3}. It is calculated by verifying the data type of each column to its values. To measure uniformity, we count the number of cells where the data type does not match the column and divide the result by the total number of cells in the dataset. We take the mean of these incorrect matchings over the number of columns in a dataset to generate the final percentage of uniformity. 
\subsubsection{Metadata coupling}\label{Metadata coupling}    
    The current empirical study includes a unique highlighted novel approach for including metadata information along with the dataset. Metadata refers to structured information provided along with the dataset describing its columns and respective values. It helps the user gain more insight into the dataset and understand the relationship between data and columns. \textbf{Metadata coupling} checks the metadata matching score of data by comparing the column name to the column description provided by the metadata codebook for any incoming dataset \cite{metdatanew}. Byrne et.al \cite{3} described that \enquote{adherence to metadata standards is an aspect of Data Quality}. Metadata should comply with the dataset columns and clearly define its purpose. We approached this problem with a classical view of natural language processing and combined character based, token based, feature based and phonetic based similarity algorithms.The below mentioned points refer to the four types of 13 similarity algorithms illustrated in the current study. 
    \begin{enumerate}
        \item \textit{Character based similarity} included in our algorithm are Hamming distance \cite{t1}, Levenshtein distance \cite{t2}, Jaro-winkler distance \cite{t3}, Needleman wunsch \cite{t4}, Smith waterman \cite{t5} and longest common subsequence \cite{t6}. It is also known as edit distance measure, takes two strings and calculates the edit distance that is the minimum number of edits required to transform one string into the other. It is useful in recognizing typographical errors \cite{metadata3}.

        \item \textit{Token based similarity} models include Jaccard similarity \cite{t8}, Cosine similarity \cite{t9}, Manhattan distance \cite{t10},Tanimoto similarity \cite{t12}. They encompass situations where each string in the sentence is a set of tokens and similarity is calculated by manipulation of these tokens \cite{t7}. The similarity is greater if there is an overlap of tokens in the two matching sentences. 

        \item \textit{Feature based similarity} uses set theory operations between features to calculate sentence similarity. We included the Tversky similarity \cite{t11} and the overlap algorithm in our model \cite{metadata3} to capture the essence of the similarity with respect to the sets defined by matching algorithms.

        \item \textit{Phonetic based similarity} approaches like match rating uses variation of sound to recognize misspelled data to calculate sentence similarity \cite{metdata4}.
    \end{enumerate}

    We combined the above mentioned 13 algorithms to create a hybrid approach to convert the abstract term of metadata matching to a measurable quantity. The process of calculation of the matching score consisted of three steps. 
    \begin{itemize}
        \item Firstly, data preprocessing and vectorisation, which includes conversion into lowercase, removal of stop words and symbols, stemming and lemmatization\cite{metadata2}.
        \item Further these sentences are converted into feature vectors which aid in the calculation of similarity score of all the algorithms separately.
        \item Vectors generated in the first step are fed into these algorithms for similarity calculation.
        \item The similarity scores from these string comparison algorithms are normalized to values between zero and one to make them comparable amongst the 13 defined algorithms. After normalization, the score zero represents low similarity, and one represents high similarity. 
    \end{itemize}    
    The final step is calculation of metadata matching score. With equal weightage to each algorithm, this score averaged over thirteen helps generate Metadata Coupling percentage allows to adjudge the concordance of the incoming dataset with the descriptive codebook and ultimately provides a measurable compliance score for the same. 
\subsubsection{Statistics}\label{Statistics}    
     \textbf{Statistics} modelling of a dataset is performed by calculating and studying the percentage of missing cells, duplicate rows, and skewness. The total percentage of non-missing cells and non-duplicate rows are calculated to aid in its quality judgement. 
    \begin{itemize}
        \item \textbf{Missing Cells}: As stated by Jayawardene et. al. \cite{3} \enquote{Data is complete if no piece of information is missing}. Large portions of missing cells therefore, reflect poor data quality and render the dataset useless. 
        \item \textbf{Duplication of Values}: Sidi et. al. \cite{6} stated that a measure of unwanted duplication within the dataset indicates that the respective data is inappropriate for usage among end users as it contains redundant data. 
        \item \textbf{Skewness}: Skewness is a measure of the lack of symmetry in a dataset distribution. Any symmetric data should have a skewness nearing to zero, whereas, data that does not include objectivity would be highly skewed. The skewness of the dataset helps the user understand if any bias is present in the data. Highly skewed data would represent unfair statistics and would not yield good results when applying that dataset. A symmetric or unbiased dataset/survey would always have zero skewness, and hence the lower value of skewness would indicate a higher percentage of data quality \cite{4}. 

    \end{itemize}
    
\subsubsection{Correlations}\label{Correlations}    
    \textbf{Correlations} between columns of a dataset is a way of understanding the relationship between its multiple dimensions or features. We use Pearson's correlation coefficient to generate a percentage of correlation of the dataset wherein the high value would indicate low data quality and vice versa. The system finds the highly correlated columns and presents them to the user in the detailed report. High correlation means a noisy dataset, which can either be helpful in certain situations or harmful in some. Hence, it's up to the user to decide and generate the score for the same. We provide a comprehensive report on interrelationships between variables used to guide the user if these variables need to be removed or kept.

\subsection{Metric Formulation}\label{metric formulation}
Our system uses the above mentioned “ingredients” in section \ref{introduction} to formulate a metric that further carefully analyses and compares data quality trends on the testing datasets.

Principal Component Analysis or PCA is used to assign weights to input variables and generate innovative indices. We create data-driven indices by aggregating input variables from our training data by using PCA loadings \cite{pca}. In this approach, we formulated coefficients of the linear combination of the original variables from which the principal loadings are constructed. These loadings can be both positive and negative; wherein, positive loadings indicate a positive correlation between the variable and the principal component, and negative loadings indicate a negative correlation. Large (either positive or negative) loadings suggest that a variable has a strong effect on that principal component. These loadings to signify to formulate the percentages for each ingredient towards the DQ score. However, judging the negative loadings posed a grave issue. Hence, we shifted the offset of these values from [-1,1] to [0,2] by adding 1 to all principal component loadings to make all values positive. Furthermore, these loadings are normalized to retrieve the percentage of each “ingredient” over a total of 100 as shown in table \ref{tab:pca loadings}. The first column in the table \ref{tab:pca loadings} contains PCA loadings which are both positive and negative. The second column contains the values when the offset is shifted to make all loadings positive. Further, normalization is done by dividing the value of each loadings by the sum of all the loadings of all ingredients to get the percentage of each ingredient over the DQ score. 

The above approach formulated the metric (figure \ref{fig: label}) that was tested to calculate data quality scores of the Demographics and Health Surveys (DHS) Program Indian dataset over the years 1992-93, 1998-99, 2005-06 and 2015-16.

    \begin{table}
    \centering
    \resizebox{\linewidth}{!}{%
    \begin{tabular}{|l|l|l|l|l|}
    \hline
    Labels                & Principal Component Loadings & Positive Principal Component Loadings & Normalization & Percentage  \\ \hline
    Provenance            & 0.066                        & 1.066                                 & 0.097032587   & 9.703258693 \\ \hline
    Uniformity            & 0.867                        & 1.867                                 & 0.169943565   & 16.99435645 \\ \hline
    Dataset Characterstics & 0.87                         & 1.87                                  & 0.170216639   & 17.02166394 \\ \hline
    Metadata Coupling      & -0.059                       & 0.941                                 & 0.085654469   & 8.565446932 \\ \hline
    Non-Duplicate Rows      & -0.205                       & 0.795                                 & 0.072364828   & 7.236482796 \\ \hline
    Non-Missing Rows        & 0.108                        & 1.108                                 & 0.100855634   & 10.08556344 \\ \hline
    Un-skewness            & 0.702                        & 1.702                                 & 0.154924449   & 15.49244493 \\ \hline
    Inconsistent Categorical Columns    & -0.085                       & 0.915                                 & 0.083287821   & 8.328782086 \\ \hline
    Un-correlation         & -0.278                       & 0.722                                 & 0.065720007   & 6.572000728 \\ \hline
    TOTAL                 &                              & 10.986                                &               & 100         \\ \hline
    \end{tabular}}
    \caption{Principal Component Loadings}
    \label{tab:pca loadings}
    \end{table}

\section{Results}\label{results}
The underlying study encompasses the ingredients of data quality and provides a quantitative metric to measure it. In this section, we showcase the DQ label, a case study using the proposed metric, the validation methodology and the created automated data quality platform. We aim to use a case study involving DHS datasets of India for the years 1998-99, 2005-06 and 2015-16 and various synthetically modified datasets from Kaggle. We have also created a data quality platform that enables users to utilize the benefits of the proposed metric and retrieve a comprehensive report containing values of these defined ingredients. The report also provides ways and means to improve the quality of data to the researcher. 

\subsection{Data Quality Label}
\begin{figure}
        \centering
        \includegraphics[height=10cm, width=8cm]{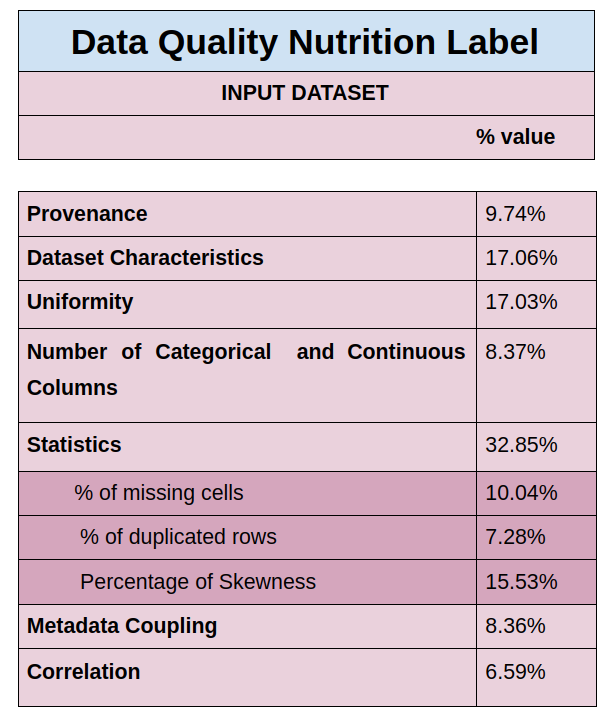}
        \caption{Quality Metric: Data Quality Label}
        \label{fig: label}
    \end{figure}

Using the above mentioned procedure to get the PCA metric in section \ref{metric formulation}, we get the following label as shown in figure \ref{fig: label}. This label provides percentage weights for all ingredients discussed in this study and finally adds all weights upto 100\% out of which the DQ score is calculated. 

\subsection{Case Study: DHS Indian Datasets}
This sections highlights the case study carried out by the current research to illustrate the above proposed metric and show data quality trends. The Demographic and Health Surveys (DHS) are a nationally representative household survey program that provide data for a wide range of monitoring and impact evaluation indicators in population, health, and nutrition \cite{dhs3}. According to the recode manual for each respective year of the DHS program, there are a total of 17 sections. Namely, \textit{Respondent's basic data, Reproduction and Birth History, Reproduction, Contraceptive Table +Contraceptive Use, Maternity, Maternity and Feeding, Health History along with Height and Weight, Marriage, Fertility Preferences, Partner's Characteristics and Women's Work, AIDS and Condom Use, AIDS, STIs and Condom Use continuation, Calendar, Maternal Mortality, Malaria, and Domestic Violence}. 

For a long time, the Demographic Health Survey (DHS) project was considered the gold standard for nationally represented data collection \cite{dhs6}. Heavy emphasis on data quality is a hallmark of DHS surveys, and hence computer-assisted personal interviewing (CAPI) is used by DHS officials to improve data quality \cite{dhs5}. The data collection process is supervised on every level. The team supervisors and field editors provide the first level of supervision. The second level of supervision consists of staff visits to the field. These supervisors are responsible for closely monitoring the teams' work to ensure that all sampled households are visited, and all eligible respondents are contacted. All the factors mentioned above ensure the high data quality of the DHS dataset \cite{dhs5}.

As an application to the metric proposed, we conducted an in-depth study on the DHS India Dataset (Individual Recode) for 1998-99, 2005-06 and 2015-16. The data collected from surveys conducted in these three years was divided into sections based on the division on the DHS Recode Manual, commonly called the codebook. All these divided sections were considered independent datasets and analyzed for data quality using the proposed quality metric. The DHS program has three years of data collection consisting of 17 sections, combining a total of 51 testing datasets.

\begin{figure}
        \centering
        \includegraphics[height=8cm, width=13cm]{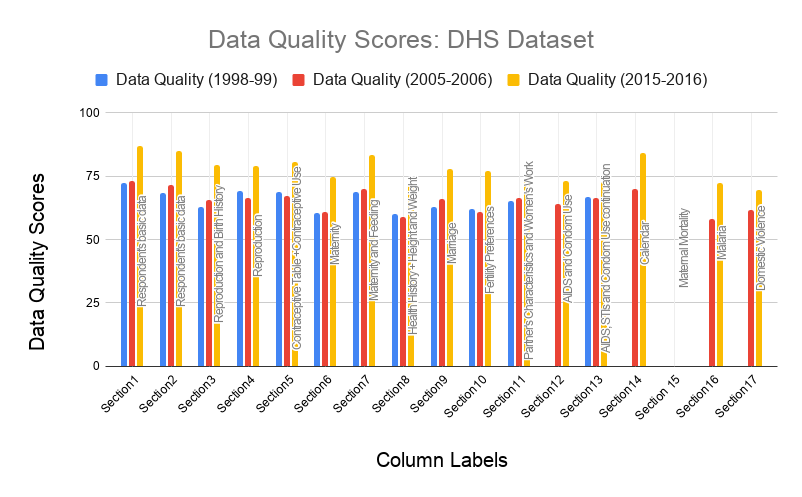}
        \caption{Data Quality Scores over the years divided among sections: DHS India Dataset}
        \label{fig: Data Quality Scores: DHS India Dataset}
    \end{figure}

The DHS data cites the provenance details, including the source country, origin on its website, and data processing details have been specified in the survey org manual \cite{dhs6}. All the datasets follow the rules of uniformity and dataset characteristics according to the proposed metric as the values of mean, median, mode, and type of data match accordingly. After conducting statistical analysis on all the datasets, the model observed that all the values match perfectly with the data provided by DHS recode manual. Hence, the value of provenance, uniformity and data characteristics are 100 for all sections of the datasets from the three years. The metadata codebook supplied with the dataset was used to calculate the text-similarity scores with the column descriptions in the recode manual to calculate the metadata concordance scores. The recode manual for all the years is similar except for new sections and changes in columns of existing sections. After in-depth analysis, we observe that the metadata matching score ranges from 85-95\% for each section, with a few exceptions. While observing the results, we see a constant increase in the metadata matching scores over the years in almost all sections, which support the hypothesis that individual variables of metadata coupling improve with time and technology. The comprehensive report provides the details of correlations between all the columns from highest to most minor correlation. This is calculated using the Pearson correlation coefficient and provides us with names of the highly correlated columns if their coefficient values are greater than 0.8. In the DHS dataset, we observe varying correlations ranging from 2\% to 40\%. An in-depth analysis of missing cells, duplicate rows and skewness were carried out, which showed that the percentage of missing cells range from as low as 8.23\% to as high as 34.2\%. The percentage of duplicate values in most sections is almost equal to zero as the testing dataset is a survey dataset which means that each row is a unique individual. For any researcher, categorical datasets are easier to read and have fewer chances of being misinterpreted. Hence, the datasets were checked if in the columns of categorical data, the field worker by mistake entered continuous data which would result in bad analysis results and hence bad data quality. Further, we calculated the number of cases where the columns were defined as categorical in the recode manual but were found to be continuous in the dataset and vice-versa to get the values of inconsistencies in their description. These values were divided by the total number of columns in that data to provide an inconsistency score for the study. Most sections showed low fractional values which reflected good data quality results. The results were recorded and displayed on the platform developed and in the comprehensive report. 
After calculating the scores for all data quality ingredients and calculating the data quality for each section for the years 1998-99, 2005-06 and 2015-16, we observe an increase in data quality as seen in figure \ref{fig: Data Quality Scores: DHS India Dataset}. Each bar here refers to the DQ score calculated using the proposed metric. This score is grouped over the years of the DHS collection program to compare how the quality is changing with time. The figure clearly sheds light on the hypothesis that DQ is increasing with the passage of time and increasing technology. It provides us with better visualization of the increase and decrease of data quality over three consecutive surveys. 

Our analysis shows that the DHS dataset has a high data quality which is in concordance that data quality is the primary factor for creating survey DHS datasets.

\subsection{Validation of Metric using the Principles of Mutation Testing}
Mutation testing is an error-based testing technique involving the construction of test data designed to uncover specific errors in metrics \cite{mutationtesting}. Inspired by this idea, the current empirical study implemented a modified version of mutation testing by forming test data after many mutated versions from the original version. We formed the idea of Mutation Testing wherein some aspects of data are changed, and noise was added to check if our metric can identify these errors. The goal of Mutation testing in data quality is to ensure the authenticity of the defined metric and prove whether it can capture the essence of data quality. Synthetic Datasets are noise-induced datasets generated through computer programs having some form of corruptness. If the proposed metric in this study successfully detects the percentage of noise included by the researcher, then we can say that the metric ultimately measures the quality of data. 

In this research, we generated synthetic datasets from existing healthcare datasets from Kaggle by intentionally introducing and removing impurities.
\begin{enumerate}
    \item Removing Impurity here refers to factors that would increase the data quality of a dataset. Examples include removing missing cells and duplicate rows, eliminating columns with high correlation, or making the metadata/codebook more descriptive. 
    
    \item Introducing Noise highlights the factors that would reduce data quality by making it unfit for machine learning/artificial intelligence algorithms. Examples include adding missing cells or duplicated rows, changing the format of a few cell components making it less uniform/ machine-readable or deteriorating the metadata. It does not convey the meaning of a column heading accurately.
\end{enumerate}

\begin{figure}
        \centering
        \includegraphics[height=6cm, width=13cm]{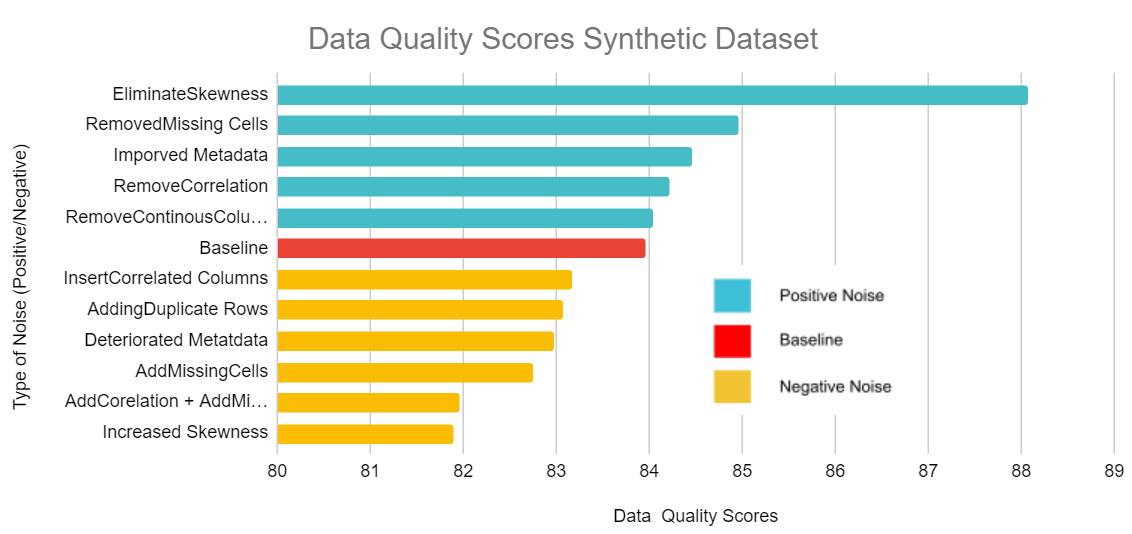}
        \caption{Variation in Data Quality Scores on Synthetic Datasets}
        \label{fig: Data Quality Scores Synthetic Dataset}
    \end{figure}

Synthetic datasets were created by adding a combination of positive and negative noise to original data. For every existing dataset, ten additional datasets were analyzed to formulate results for our study.  The trends are shown in figure \ref{fig: Data Quality Scores Synthetic Dataset} wherein if the noise is added, i.e., adding missing cells and duplicate rows and adding columns that increase correlation, the data quality gets reduced. However, when the study eliminates the missing cells and duplicate rows, removes all skewness in the dataset and improves the metadata description for all columns, the data quality is improved. The baseline in the given figure as the red bar refers to the original, unmutated dataset, whereas the green refers to the improvement in data quality and yellow bars refers to the deterioration in data quality. 

This Mutation Testing of datasets shows that the metric proposed in the underlying study correctly captures the quality of data which would help the researcher gain insights before using it further for research purposes. It would eliminate any instances where a machine learning/artificial intelligence model fails and provides a foundational concept in the study of data quality.

\subsection{Data Quality Platform}
Machine learning and Artificial intelligence models are severely dependent on the quality of data. Erroneous decisions and results resulting from bad data are inconvenient and time-consuming and, many times, costly. In this study, we also present a platform to accurately measure data quality using the proposed metric that provides integrated statistical and visual analysis to summarise the quality of a dataset. It enables the users to generate a comprehensive report highlighting the problems with their dataset and presents ways and means to improve this quality index. 

\begin{figure}
    \includegraphics[height=8cm, width=13cm]{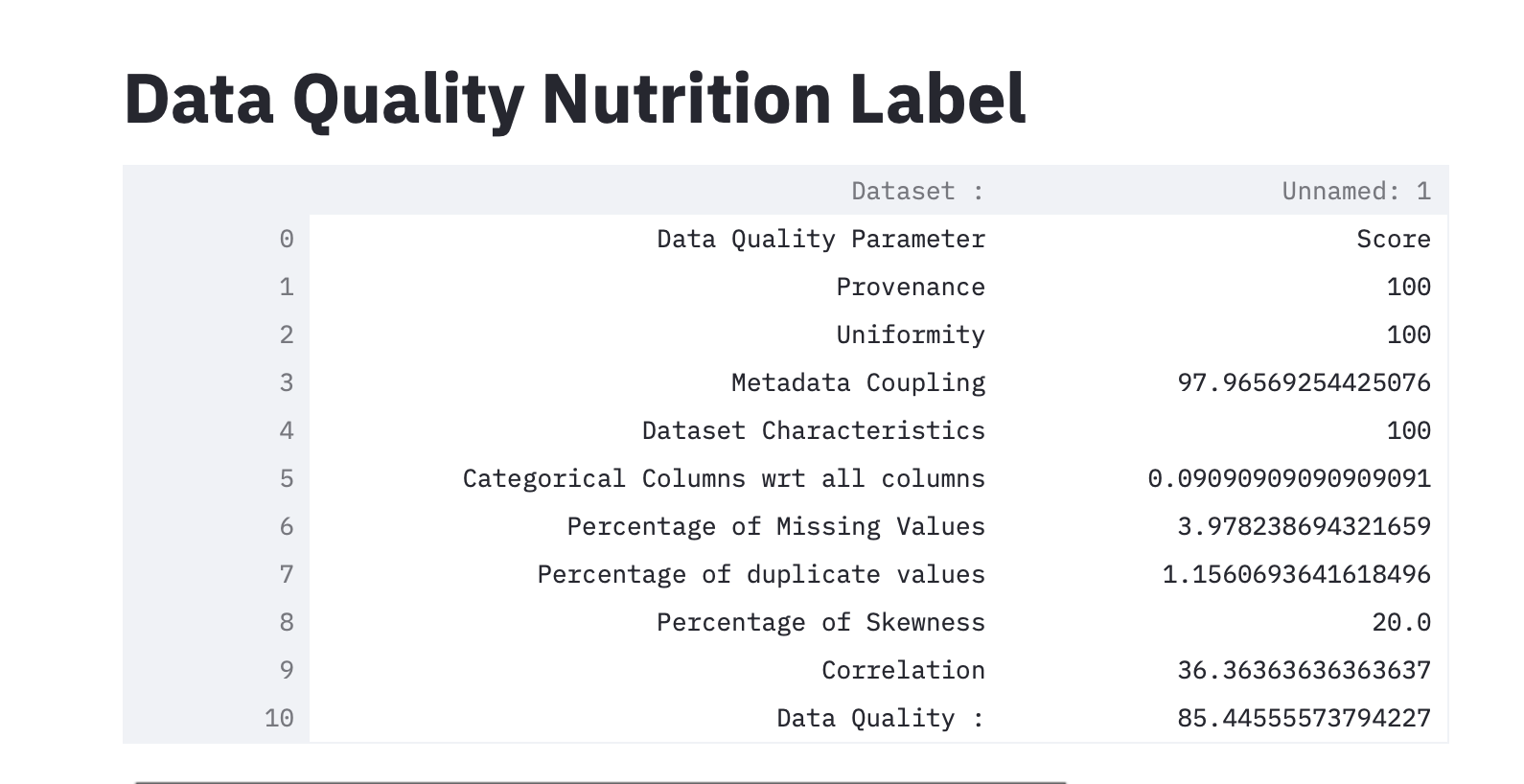}
    \caption{Data Quality Platform : Quality Label Display}
    \label{fig: platform3}
\end{figure}

The platforms take an incoming dataset and the metadata file in an SPSS, CSV or XLS format as input.  Further, it performs any preprocessing if required as mentioned in the codebooks. Further, this data is fed into our quality model which analyses all data quality parameters and evaluates the data quality score as shown in figure \ref{fig: platform3}. Our dynamic platform enables the user to select any variable/column and see all its descriptive characteristics, including the data type, mean, median, maximum and minimum value. The correlation graph of the dataset can also be viewed using a simple check box. We present the user with the names of the columns that are highly correlated for future analysis. The values of the data quality parameters as mentioned in the section \ref{introduction} are displayed on the dashboard. The model uses these measured values to fill the quality label which is further added to the comprehensive report. This comprehensive report includes details of all the ingredients and mentions columns where skewness and correlation can be decreased, if any. It also shows columns where the model received low metadata coupling values and suggests improving the description of those columns. It highlights the missing values rows by green and duplicate values rows by yellow to enable interactiveness with the end user. 

If the user wants to learn about the metric and its parameters, they can navigate the “About the metric” section from the drop-down menu and select the parameter to learn more. On choosing the data quality label section, the platform displays the value of all the parameters in the form of a label, as shown in figure \ref{fig: label}. The data quality platform is available on the cited link for reference \cite{platform}.

\begin{figure}
\centering
    \begin{minipage}{.5\textwidth}
        \centering
        \includegraphics[height=7cm, width=6cm]{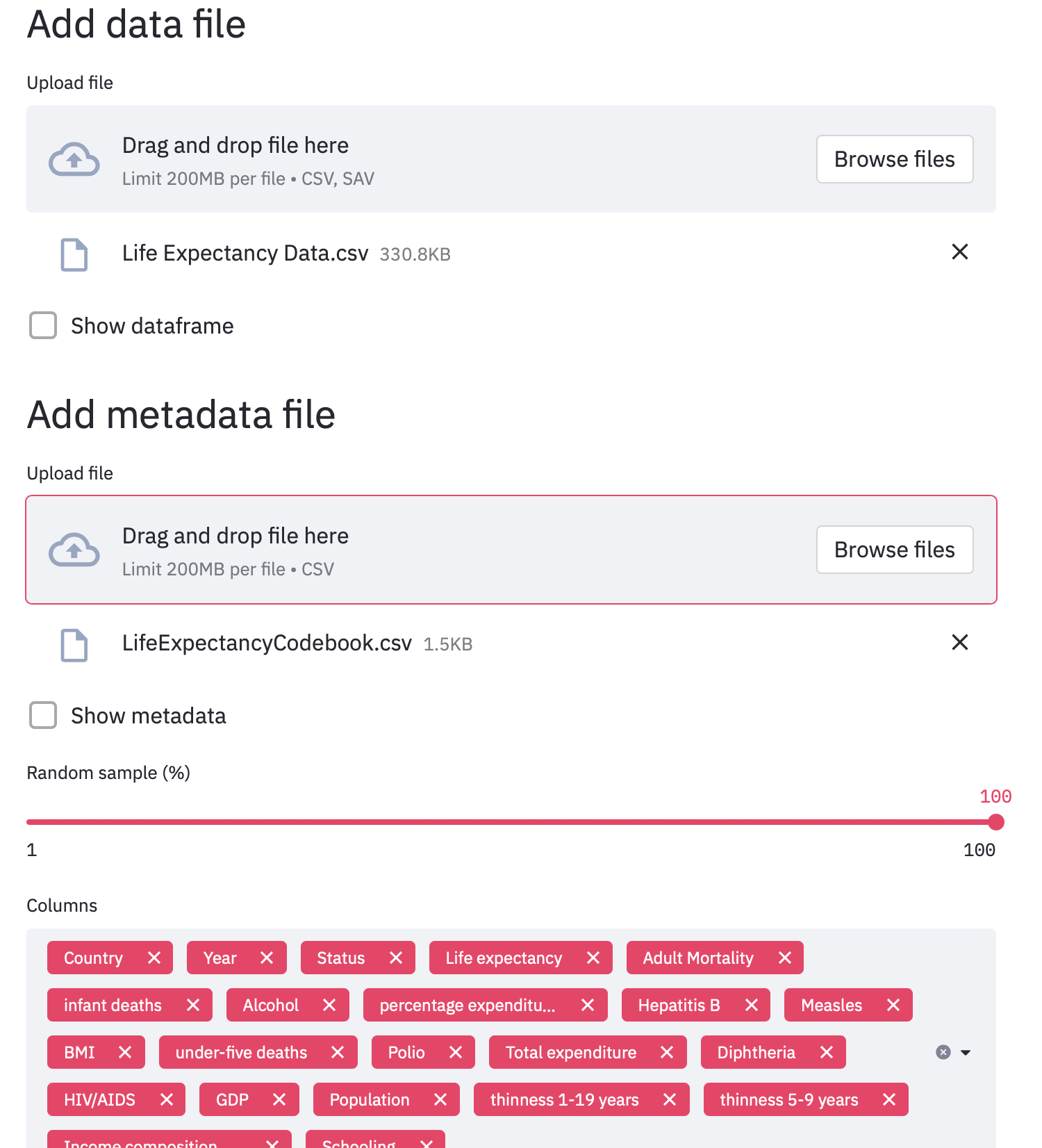}
        \caption{Data Quality Platform : Data Explorer Tab}
        \label{fig: platform1}
    \end{minipage}%
    \begin{minipage}{.5\textwidth}
        \centering
        \includegraphics[height=7cm, width=8cm]{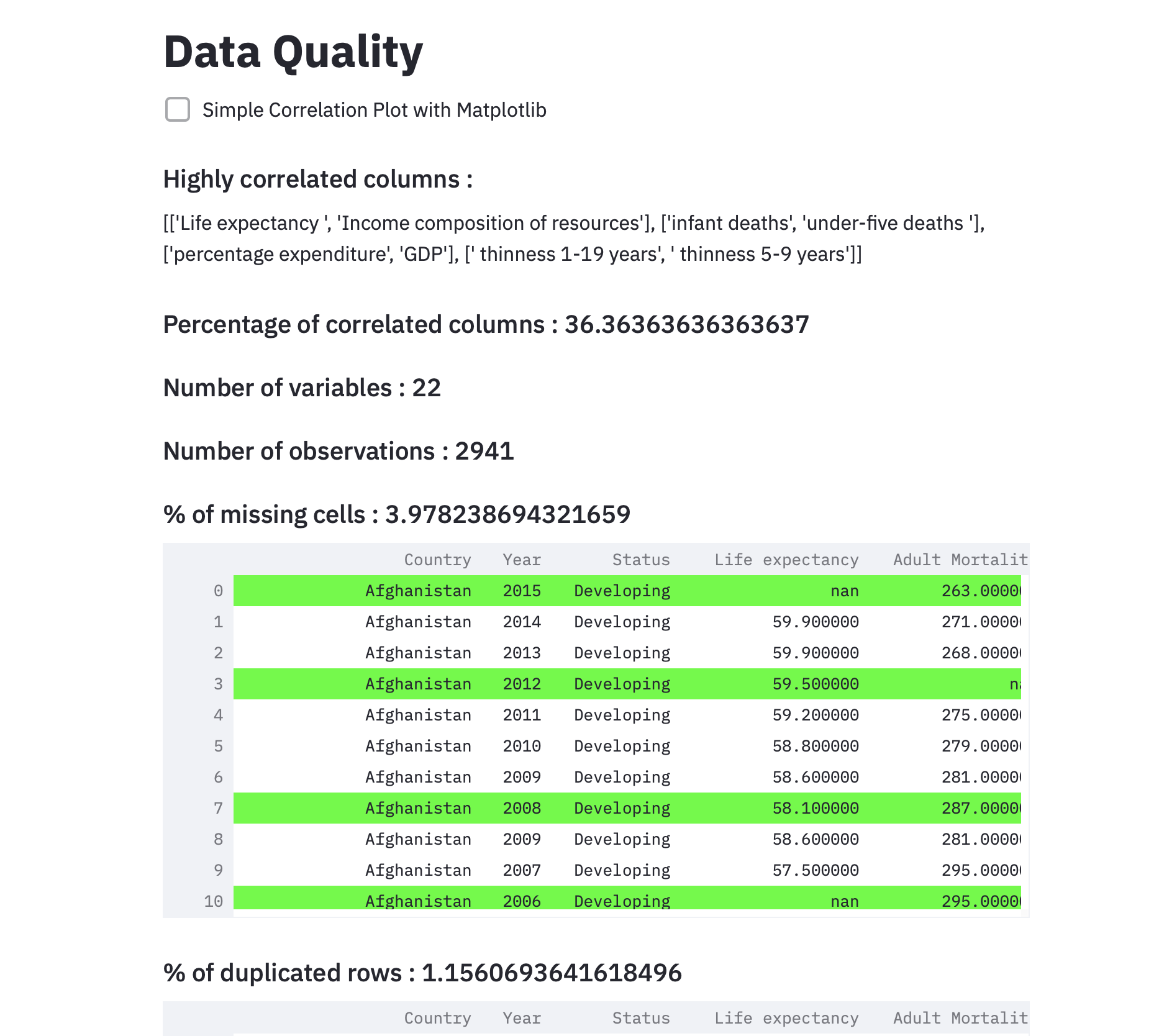}
        \caption{Data Quality Platform : Data Quality Parameters}
        \label{fig: platform2}
    \end{minipage}
\end{figure}

\section{Conclusion and Future Work}
\begin{displayquote}
You can't control what you can't measure
\\
- Tom DeMarco
\end{displayquote}
Increased use of data has urged the need for quality data for decision making. Hence data quality checks and their interpretation has become the need of the hour. This can be achieved when the state of the art technologies come into existence to improve the data quality. This includes the coupling of carefully analyzed and discussed \textbf{Data Quality “ingredients”} to further improve upon the quality of a dataset.
Following the words of Tom DeMarco, we aimed to quantify data quality and formulate an approach to measure the same with and aim to improve it further.

In an effort to improve the current state of practice of data analysis, in this research study, we created the \textbf{Dataset Nutrition Label}, a diagnostic framework that provides a concise yet robust and standardized view of the core components of a dataset. Assessing data quality is an on-going effort that requires awareness of the fundamental principles underlying the development of subjective and objective data quality metrics. In our research, we represent subjective and objective assessments of data quality in terms of scores generated that check the quality of data. We have developed illustrative metrics for important data quality dimensions.

Finally, we have presented an approach that combines the \textit{subjective and objective assessments} of data quality and demonstrated how the approach can be used effectively in practice. Together, this provides \textit{flexibility, scalability, and adaptability}. With this approach, data specialists can efficiently compare, select, and interrogate datasets. They can provide qualitative and quantitative modules that leverage different statistical and probabilistic models.  As a result, data specialists have a better, more efficient process of data interrogation, which will produce efficient Artificial Intelligence models. This research could be the first step in a broader effort toward improving the outcomes of Artificial Intelligence systems that play an increasingly central role in our lives.

With advancements in technology and the creation of new systems every day, data collection and processing have improved immensely. As an application of our proposed metric, an in-depth case study was performed on DHS data for India (Individual Recode)for 1998-99, 2005-06, and 2015-16. The datasets were divided into sections based on the recode manual and thoroughly tested on the proposed nine data quality ingredients. The scores for all data quality ingredients and the data quality for each section over 1998-99,  2005-06, and  2015-16 were calculated. We observed that the score for provenance and uniformity is 100 for all three years. The metadata matching score ranges from 85-95 for each section; it also displayed a constant increase over the years. The score for other ingredients like missing cells, duplicated rows, skewness, correlation and categorical column inconsistencies showed varying results but improved over the years in almost all sections. DHS dataset is considered the gold standard for nationally represented data collection, which was in concordance with the results of our case study.

To further test if the proposed metric can accurately capture various datasets with different data quality issues, we performed mutation testing. For this research, we generated synthetic datasets from existing healthcare datasets from Kaggle by intentionally introducing and removing impurities. After the creation of these synthetic datasets, data quality for each was calculated. We observed that eliminating impurities from the dataset increased the data quality while adding impurities decreased the data quality compared to the baseline dataset. This observation shows that the metric proposed in the underlying study correctly captures the change in data quality accurately and can help the researcher gain insights before utilizing the data for research.

Quality of data is an ever growing aspect, we can never stop increasing data quality. In our research, we formulated a metric and a data quality platform which can we used by any user to formulate a score of their dataset and utilize it in the best possible way.In the future, we plan to elaborate our metric by improving metadata matching algorithm using sentimental word importance and other corpus, knowledge and hybrid based text similarity algorithms. We also aim to refine the platform by incorporating datasets in forms other than CSV or SPSS and a feature can be added to read metadata directly from the website or from the code book which is in the form of a PDF. Along with refining the platform we aim to automate the process of calculation of ingredients like provenance and uniformity. The platform will be made more user friendly and more visualization techniques can be added to help the researcher study data in a better way. In addition to this there is also a scope of expanding this metric to datasets of various formats in and outside healthcare. Additional information highlighting the problematic areas of a dataset along with suggestions to improve the data quality can be provided to the owner of the dataset/survey members. We feel that after all these improvements, our project will be well enough for deployment.

\bibliography{Biblography}

\end{document}